\documentclass[letterpaper, 10 pt, conference]{ieeeconf}  
\pdfoutput=1

\IEEEoverridecommandlockouts                              

\usepackage[dvipdfmx]{graphicx}
\usepackage{amsmath}
\usepackage{amssymb}
\usepackage{algorithm, algpseudocode}
\usepackage{times}
\usepackage{url}
\usepackage{color}
\usepackage{CJKutf8}

\newcommand{\figref}[1]{Fig.~\ref{figure:#1}}
\newcommand{\tabref}[1]{Table~\ref{table:#1}}

\DeclareMathOperator{\sgn}{sgn}
\DeclareMathOperator{\atan2}{atan2}

\title{\LARGE \bf
Foundation Model based Open Vocabulary Task Planning and Executive System for General Purpose Service Robots
}                               

\author{Yoshiki Obinata$^{1}$, Naoaki Kanazawa$^{1}$, Kento Kawaharazuka$^{1}$,\\
  Iori Yanokura$^{1}$, Soonhyo Kim$^{1}$, Kei Okada$^{1}$ and Masayuki Inaba$^{1}$
  \thanks{$^{1}$The authors are with the Graduate School of Information Science and Technology, The University of Tokyo, 7-3-1 Hongo, Bunkyo-ku, Tokyo, 113-8656, Japan. [obinata, kanazawa, kawaharazuka, yanokura, s-kim, k-okada, inaba]@jsk.imi.i.u-tokyo.ac.jp%
  }
}

\begin{document}

\maketitle
\thispagestyle{empty}
\pagestyle{empty}

\begin{abstract}
This paper describes a strategy for implementing a robotic system capable of performing General Purpose Service Robot (GPSR) tasks in robocup@home. The GPSR task is that a real robot hears a variety of commands in spoken language and executes a task in a daily life environment. To achieve the task, we integrate foundation models based inference system and a state machine task executable. The foundation models plan the task and detect objects with open vocabulary, and a state machine task executable manages each robot's actions. This system works stable, and we took first place in the RoboCup@home Japan Open 2022's GPSR with 130 points, more than 85 points ahead of the other teams.


\end{abstract}

\section{Introduction}
\label{introduction}
Developing a general purpose daily life support robot system is the dream of humankind. In the future, when an aging society is expected to arrive, the demand for daily life support by robots will increase at home \cite{yamazaki2012home}.

There are mainly two problems for daily life support robots to work at home, understanding spoken language and generate life support actions. Especially in generating life support actions, there are various researches for robots to acquire a general knowledge of the human world to generate sequences of daily life support actions, such as observing the daily life environment \cite{tenorth2009knowrob} and using web data \cite{tenorth2011web}.

In recent years, the prevalence of the Large Language Model (LLM) applications has enabled widespread Pre-Trained Foundation Models. It has billions of parameters and trains large data. It has shown high performance on various language tasks and has general knowledge of the world by training web data. Recently, it has been applied to learning models that handle text and images simultaneously \cite{radford2021learning}. The model that can handle text and images is called Vision Language Model (VLM).

LLMs and VLMs have been applied to robotics due to the strength of their open vocabulary task and semantic knowledge about the world. For example, it can be applied to high-level task planning \cite{ahn2022can} and real-world recognition for robot tasks \cite{kawaharazuka2023vqa}. However, the robot's actions should be decisive, and the robot must decide what to do next based on the success or failure of each action.

Based on these backgrounds, we propose a General Purpose Service Robot system that integrates an LLM, VLMs, and state machine task executable. The state machine task executable can clearly define the robot's action when it succeeds or fails, making its action decisive and stable in a real environment. We have developed a system that can perform open vocabulary robot navigation, manipulation, language tasks, and image tasks, which the state machine task executive manages. We found that our proposed system allows the robot to perform tasks given by a human in spoken language and have shown the stability of this system by getting high scores in the competition. In the future, as each recognition model and planner model is improved based on the system configuration presented in this study, we expect the robot to become more robust in understanding the state of the operating environment, manipulation, and local knowledge.

\section{System Configuration for GPSR}
\label{methods}
\begin{figure*}[tb]
  \centering
  \includegraphics[width=\linewidth]{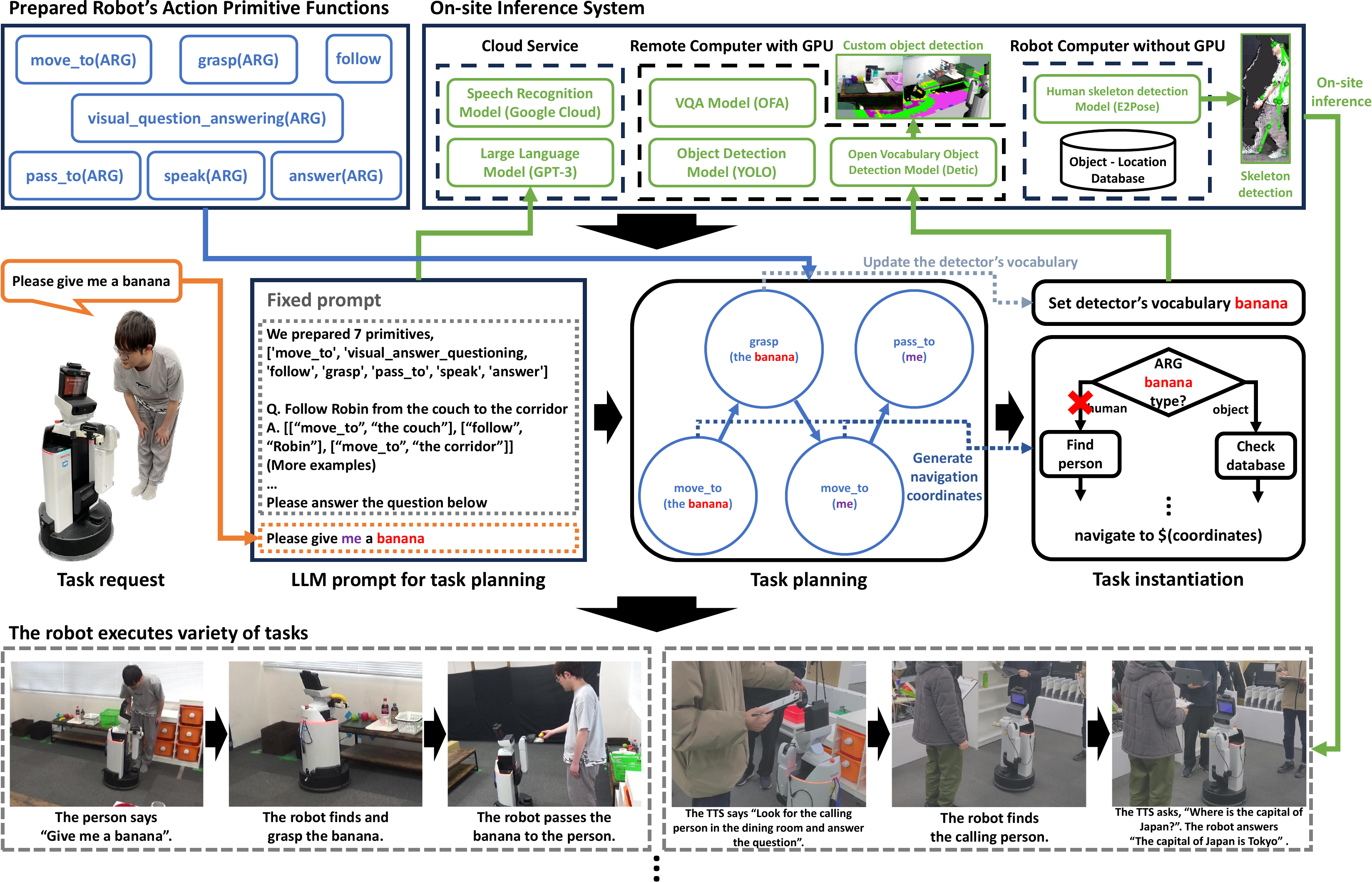}
  \caption{Configuration of the suggested GPSR system. (upper row) The robot has the following primitive functions: \textcolor{blue}{move\_to}, \textcolor{blue}{grasp}, \textcolor{blue}{follow}, \textcolor{blue}{visual\_question\_answering}, \textcolor{blue}{pass\_to}, \textcolor{blue}{speak} and \textcolor{blue}{answer}. The inference system includes a speech recognition model (Google Cloud), an LLM (GPT-3) on the cloud, a VQA model (OFA), an object detection model (YOLO), an open vocabulary object detection model (Detic) on the remote computer with GPU, human skeleton estimation model, the database where the object is a key and navigation coordinates are valued on the robot computer. (middle row) The robot hears a spoken command from a person. The robot inputs the heard command into the speech recognition model, converts it to text, and then inputs it into the LLM. The robot generates a state machine for the robot action from the output of the LLM. The robot instantiates the robot action of the generated state machine from the database or the LLM. (lower row) The robot performs variety of tasks based on the generated state machine. During the task, the robot uses the inference results of the LLM, object detection model, visual question answering model, and human skeleton estimation model.}
  \label{figure:system}
\end{figure*}

\subsection{RoboCup@home GPSR task rules}
\label{gpsrrule}
RoboCup \cite{kitano1997robocup} is a competition for intelligent robots, and there is a league called RoboCup@home \cite{wisspeintner2009robocup} that focuses on daily life support in the home. Domestic Standard Platform League (DSPL) uses an unmodified TOYOTA HSR \cite{yamamoto2018human}, and General Purpose Service Robot (GPSR) uses this HSR to understand and execute various types of task instructions from humans. We explain the GPSR rules of RoboCup@home Japan Open 2022 DSPL \cite{gpsrofficial}, in which we participated.

The main goal of GPSR is to accomplish three commands from the operator. The commands are generated using RoboCup@Home Command Generator \cite{gpsrcmdgen}, and the object's and location's name on the map are changed according to the environment of the day. The Command Generator randomly generates commands such as ``Meet William at the entrance and follow him'', ``Go to the dishwasher, meet Skyler, and take him to the entrance'' and ``Could you navigate to the storage table, look for the tray, and deliver it to me''.

First, the robot navigates to the Instruction Points, where the operator tells the commands. Then the robot executes the task. After completing the task, the robot returns to the Instruction Points to hear the command again. This procedure is repeated three times within a time limit of 10 minutes.

Regarding the score, if the robot understands each command, 10 points are added. 20, 40, and 80 points are added to operate the first, second, and third commands successfully. For each command, 25\% of the score is added if the robot moves independently to one or more appropriate locations without completely succeeding in the action, and 25\% of the score is added if the robot approaches one or more appropriate persons or objects.

\subsection{Overall System Configuration for GPSR}
To realize GPSR described in Sec. \ref{gpsrrule}, we propose an open vocabulary robot task planning and executive system that combines the foundation models and the state machine task executive. \figref{system} shows the overall system.

We prepare seven primitive functions of the robot's actions in advance. First, the operator gives a command to the robot, and then the robot inputs the command into the LLM. The LLM outputs a sequence of primitive functions and their arguments. The robot generates a state machine based on the LLM's output. If the robot needs to instantiate some primitive function with ambiguity, it queries its knowledge database or the LLM to determine detailed actions.

After the task planning, the robot executes the task based on the generated state machine. State machine task executive manages each task's actions. The robot uses the on-site inference result of the object detection model, the Visual Question Answering (VQA) model and the human skeleton estimation model.

For the inference models, we use Google Cloud Speech-to-Text for the speech recognition model, GPT-3 \cite{brown2020language} for the LLM, Detic \cite{zhou2022detecting} and YOLOv7 \cite{wang2023yolov7} for the object detection model, OFA \cite{wang2022ofa} for the VQA model, and E2Pose \cite{tobeta2022e2pose} for the skeleton estimation model. For the state machine task executive, we use SMACH \cite{bohren2010smach}.

\tabref{gpt-params} shows the GPT-3 parameters we used.

\begin{table}[h]
  \caption{GPT-3 parameters}
  \label{table:gpt-params}
  \begin{center}
    \begin{tabular}{|c|c|}
      \hline
      name & value \\
      \hline
      \hline
      model & text-davinci-003 \\
      \hline
      temperature & 0.0 \\
      \hline
      top p & 1 \\
      \hline
      max tokens & 2048 \\
      \hline
    \end{tabular}
  \end{center}
\end{table}

\subsection{Generating State Machines from a Command using Large Language Model}
\label{gensmach}
\begin{figure}[tb]
  \centering
  \includegraphics[width=\columnwidth]{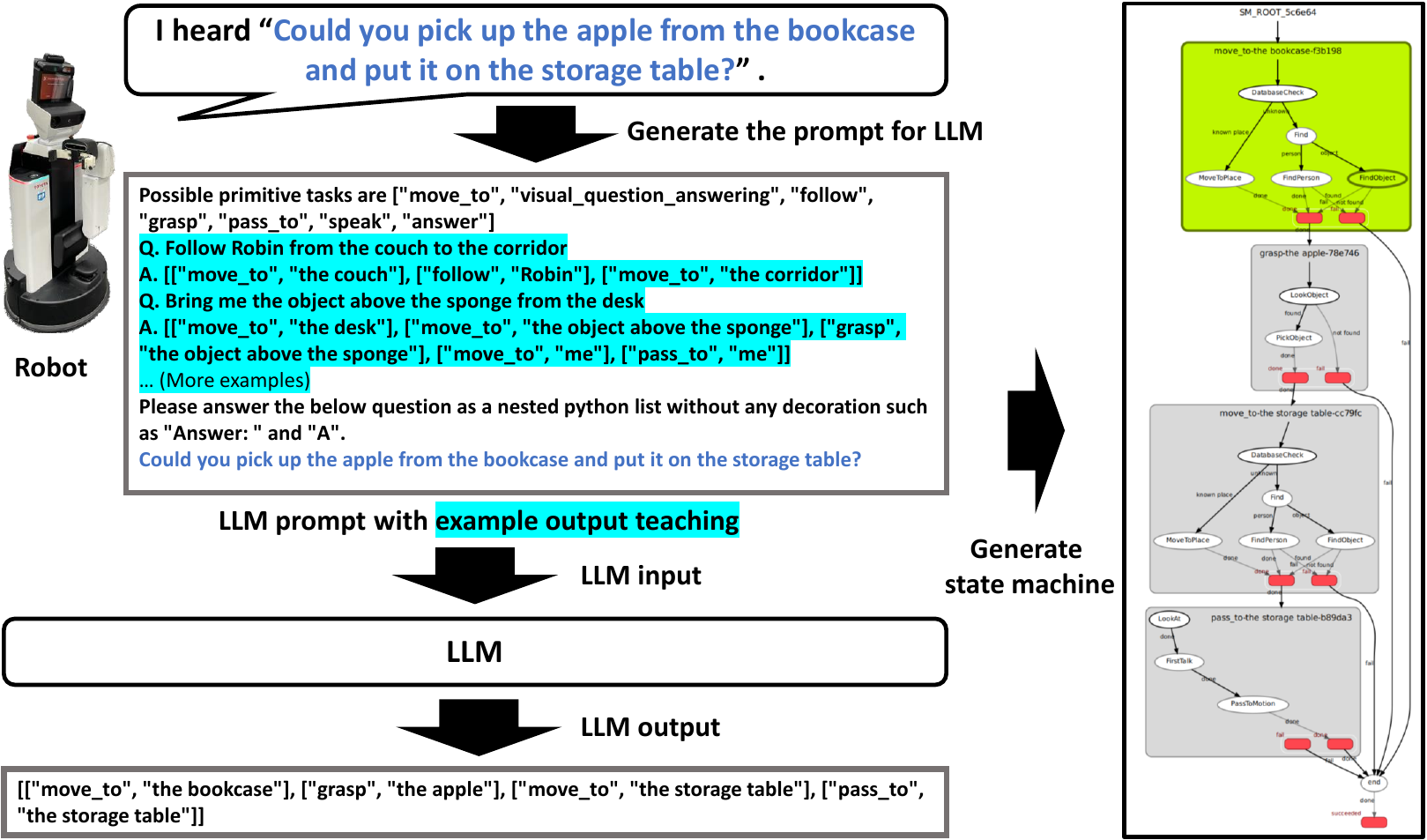}
  \caption{The process of a robot generating a state machine from a command and LLM prompts used for it. The prompt consists of the existence of seven primitive functions, examples of compiling commands to primitive function sequences, the instruction of the output format style,  and an actual command. The robot inputs it to the LLM. Then the LLM compiles the actual command into a sequence of robot actions and outputs the sequence in Python list format. The robot parses this to generate a state machine.}
  \label{figure:gen-smach}
\end{figure}

We describe a task planning method that compiles a command into primitive functions. \figref{gen-smach} shows the system. The commands are generated by RoboCup@Home Command Generator. The robot compiles these commands into the following primitive functions: \textcolor{blue}{move\_to} for the robot to move, \textcolor{blue}{visual\_question\_answering} for VQA, \textcolor{blue}{follow} for following a person, \textcolor{blue}{grasp} for grasping an object, \textcolor{blue}{pass\_to} for passing or placing an object, \textcolor{blue}{speak} for speaking to a person, and \textcolor{blue}{answer} for answering a question asked by a person. To compile a command into these primitive functions, we prepare some outputs of the Command Generator and these compilation examples and let the LLM compile a new command following the example. Specifically, enter a sentence in the LLM that consists of ``Possible primitive tasks are \textcolor{blue}{move\_to}, \textcolor{blue}{visual\_question\_answering}, \textcolor{blue}{follow}, \textcolor{blue}{grasp}, \textcolor{blue}{pass\_to}, \textcolor{blue}{speak}, \textcolor{blue}{answer}'', multiple outputs of the Command Generator and these compilation examples, ``Please answer the below question as a nested python list without
any decoration such as Answer: and A.'', and the operator's command, in that order. Then, the robot parses the output from the LLM and generates the SMACH.

Each primitive function returns the done or failure status. The robot executes the following function if a function returns the done status. If not, the robot exits the task and returns to the Instruction Point.

\subsection{Configuration of Primitives}
We describe the \textcolor{blue}{move\_to}, \textcolor{blue}{visual\_question\_answering}, \textcolor{blue}{follow}, \textcolor{blue}{grasp}, \textcolor{blue}{pass\_to}, \textcolor{blue}{speak} and \textcolor{blue}{answer} function.
\subsubsection{\textcolor{blue}{move\_to}}
\label{moveto}
\begin{figure}[tb]
  \centering
  \includegraphics[width=\columnwidth]{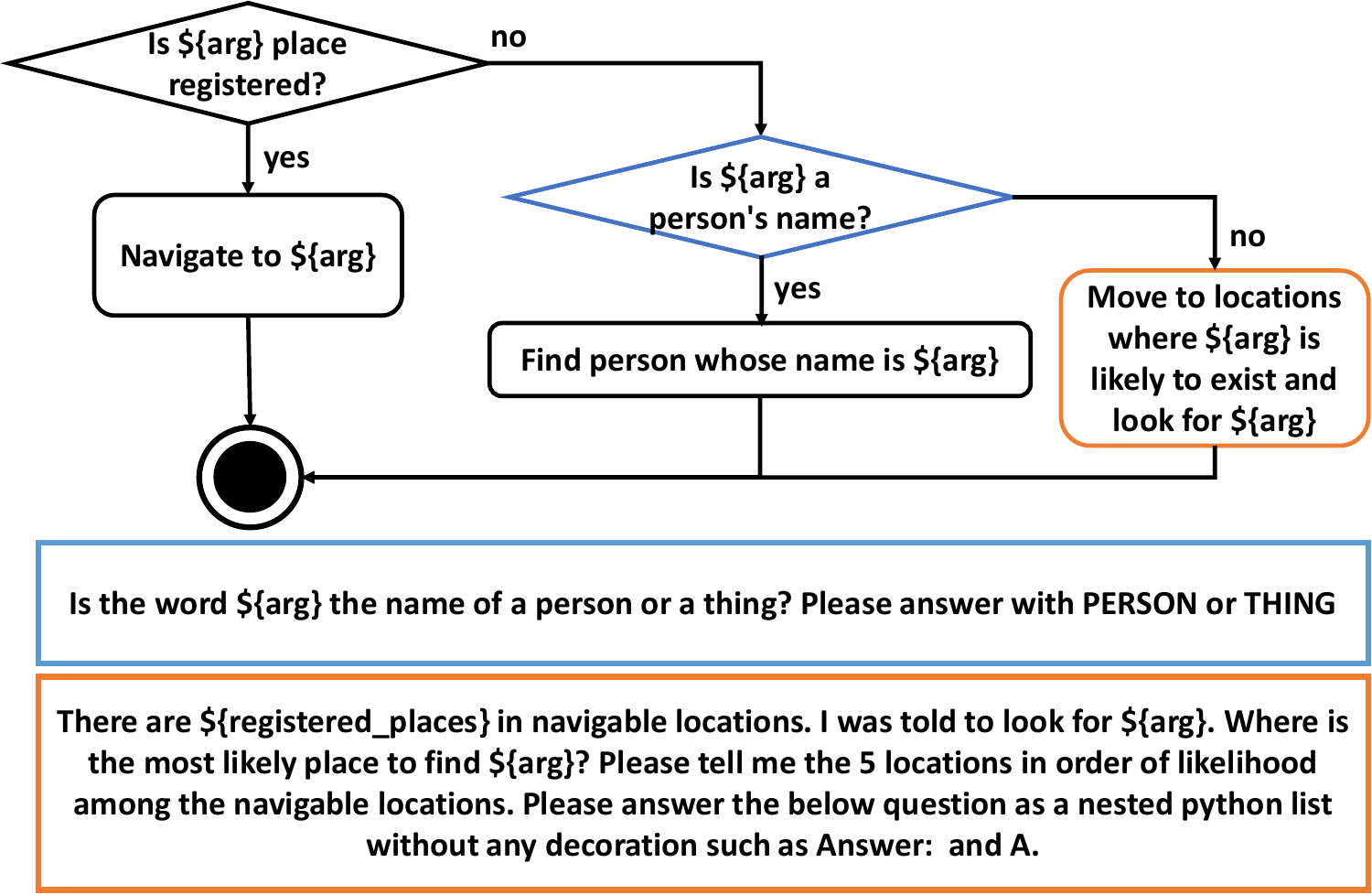}
  \caption{The \textcolor{blue}{move\_to} navigation destination determination flow and the LLM prompts. First, the robot checks if the \textcolor{blue}{move\_to} argument is a registered navigation point. If it is a registered navigation point, the robot navigates to it and exits \textcolor{blue}{move\_to}. If not, the robot determines whether the argument is a person's name; if it is a person's name, it looks for the person. If it is an object name, the robot searches for the likely location of the object from a database of known locations. The prompt in blue is a prompt to check if it is a person's name, and the prompt in orange is a prompt to suggest five possible locations for the argument's object.}
  \label{figure:move-to-gpt}
\end{figure}

We describe \textcolor{blue}{move\_to(ARG)}, which navigates to the target location. The argument may be a concrete navigation point or an ambiguous expression. \figref{move-to-gpt} shows the method for resolving this ambiguity.

If \textcolor{blue}{move\_to}'s argument is a registered navigation point, the robot navigates to it. If the argument is ambiguous, the LLM resolves it.

\begin{figure}[tb]
  \centering
  \includegraphics[width=\columnwidth]{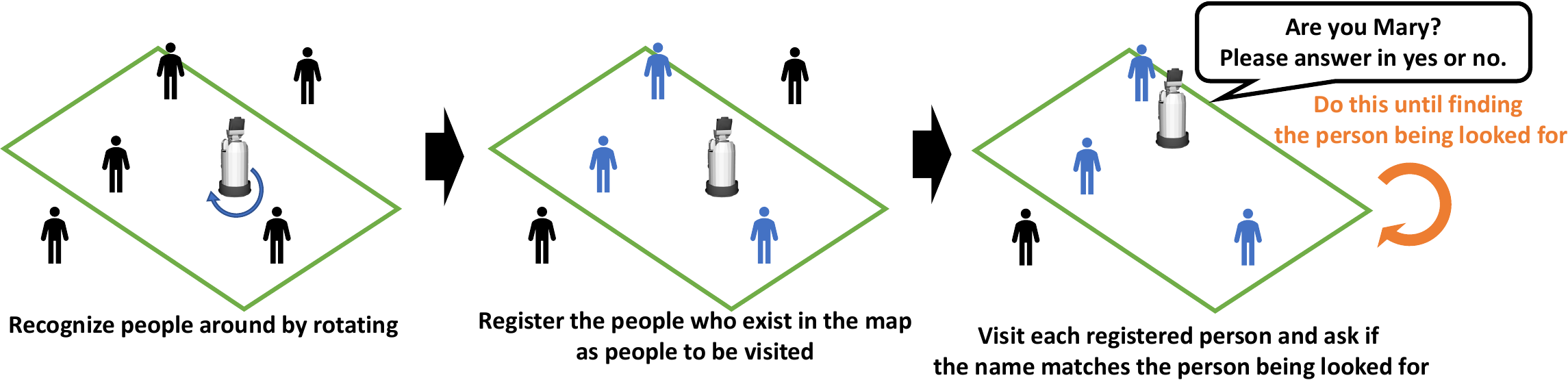}
  \caption{The action of a robot when it needs to find a person in a \textcolor{blue}{move\_to}. Using a human skeleton estimation model, the robot first rotates in place to recognize people. If the recognized person exists on the map, i.e., in the field, he/she is added to the list of people to be visited. Then, the robot visits each registered person, in turn, to check if he/she is the person it is looking for. If so, the \textcolor{blue}{move\_to} is finished; otherwise, it navigates to the next person.}
  \label{figure:find-person}
\end{figure}

We describe how the LLM resolves argument ambiguity. First, the robot checks whether the argument is a person's name. If it is a person's name, the robot finds the person of the argument, as in \figref{find-person}. The robot detects and registers people with the human skeleton estimation model while making one turn on the spot. Then, the robot visits each person detected in the competition area and asks if the person's name is the argument. If the person says ``Yes'', the robot exits \textcolor{blue}{move\_to}; otherwise, the robot visits the following person and asks the same question. This operation is repeated until the argument's person is found, and if the person cannot be found, \textcolor{blue}{move\_to} returns a failure status.

If the argument is a name of an object, the robot makes a semantic inference about where the object is likely to be at a known navigation location. For example, food is more likely to be in the kitchen, dining table, etc., and books are more likely to be on the bookshelf, and we expect the robot to navigate to those locations. The robot inputs the names of the known navigation locations and the argument into the LLM, and the LLM lists the five most likely locations. The robot visits these five locations in order. During each visit, the robot looks around itself and searches for the argument object, using the results of the object detection model. If found, the robot exits \textcolor{blue}{move\_to}; if not found, the robot navigates to the following visited location. If not found after five visits, \textcolor{blue}{move\_to} returns a failure status.

\subsubsection{\textcolor{blue}{follow}}
We describe \textcolor{blue}{follow}, which follows the person in front of the robot.

First, the robot speaks "Hi! I'll follow you. Please walk as slowly as possible to navigate me.'' and asks the follower to walk slowly. Then, it starts to follow the person.

\textcolor{blue}{Follow} assumes that a person is in front of the robot at the start of the function. While the person does not say ``arrived'' or ``over'', the robot follows the person. About the distance between the robot and the person, if it is more than 0.9 m, the robot will say ``Please come in front of the camera'' to encourage the person to come closer to the robot. If it is 0.6m - 0.9m, the robot moves straight ahead at a high speed (0.2m/s). If it is 0.5m - 0.6m, the robot moves straight ahead at a low speed (0.1m/s). If it is less than 0.5 m, the robot does not move straight ahead.

Regarding the correction of the person's angle from the robot's point of view, if the difference in the person's angle relative to the robot's direction of travel is less than 0.35 rad, the robot does not adjust the angle. If it is more than 0.35 rad, the robot adjusts by rotating at 0.3 rad/s in the person's direction. Algorithm \ref{algorithm:follow} shows the strategy of \textcolor{blue}{follow}.

\begin{algorithm}[tb]
  \scriptsize
  \caption{\textcolor{blue}{Follow} action}
  \label{algorithm:follow}
  \begin{algorithmic}
    \State{$d_{far} \leftarrow $ 0.9 m, $d_{near} \leftarrow $ 0.6 m, $d_{arrived} \leftarrow $ 0.5 m}
    \State{$\theta_{th} \leftarrow $ 0.35 rad}
    \State{$v_{fast} \leftarrow $ 0.2 m/s, $v_{slow} \leftarrow $ 0.1 m/s}
    \State{$v_{angle} \leftarrow $ 0.3 rad}
    \While{The person doesn't say ``arrived'' or ``over''}
    \State{$x, y \leftarrow$ Position of the person relative to the robot}
    \State{$\theta \leftarrow \atan2(y, x)$}
    \State{$d \leftarrow \sqrt{x^{2} + y^{2}}$}
      \If{$d > d_{far}$}
      \State{Speak ``Please come in front of the camera.''}
      \State{\textbf{continue}}
      \EndIf
      \If{$|\theta| \leq \theta_{th}$}
        \State{$v_{\theta} \leftarrow 0$}
      \Else
        \State{$v_{\theta} \leftarrow \sgn(\theta)v_{angle}$}
      \EndIf
      \If{$d \leq d_{arrived}$}
        \State{$v_{x} \leftarrow 0$}
      \ElsIf{$d < d_{near}$}
        \State{$v_{x} \leftarrow v_{slow}$}
      \Else
        \State{$v_{x} \leftarrow v_{fast}$}
      \EndIf
      \State{send $v_{x}, v_{\theta}$ velocity command to the robot}
    \EndWhile
  \end{algorithmic}
\end{algorithm}

\subsubsection{answer}
\label{answer}
We describe \textcolor{blue}{answer}, which answers the question from a person. \textcolor{blue}{Answer} is the action of asking a person if he/she has any questions, then answering after the person tells a question.

First, the robot says ``Please ask me a question''. Then the robot responds to the person's question and asks whether what it has heard is correct by saying ``yes'' or ``no''. The robot answers the question the person says ``Yes''.

The robot uses the the LLM to generate a response to a question. The robot inputs a prompt

  \begin{quote}
    My name is hsr099. I belong to JSK team.
    Current Prime ministor of Japan is Kishida. The next summer olympics would be held in Paris, France.
    Seven teams are participating in domestic standard platform league.

    Please answer the question below

    \$\{question\}
  \end{quote}

to the LLM. \$\{question\} is a spoken question. The robot speaks the output.

\subsubsection{Other Primitives}

We describe \textcolor{blue}{pass\_to(ARG)}, which passes/places the object to/at the argument. It is assumed that the robot is grasping an object. The robot speaks ``Hi! I'll give you this'', raises its arm 0.2 m, says ``Here you are!'', then opens the gripper one second after completing its speech.

We describe \textcolor{blue}{visual\_question\_answering(ARG)}, which answers the argument's question in front of the robot. The robot inputs the argument text and the image in front of it to the VQA model. Assuming the VQA models' output as \$\{answer\}, the robot says, ``I will answer the question, \$\{arg\}. \$\{answer\}'' then exit.

We describe \textcolor{blue}{grasp}, which grasps the argument's object. The robot outputs a bounding box that covers the point cloud corresponding to the region for the semantic segmentation result obtained from the object detection model. The robot approaches the end-effector to the bounding box and closes the gripper.

\section{Experiments}
\label{experiments}
\begin{figure}[tb]
  \centering
  \includegraphics[width=\linewidth]{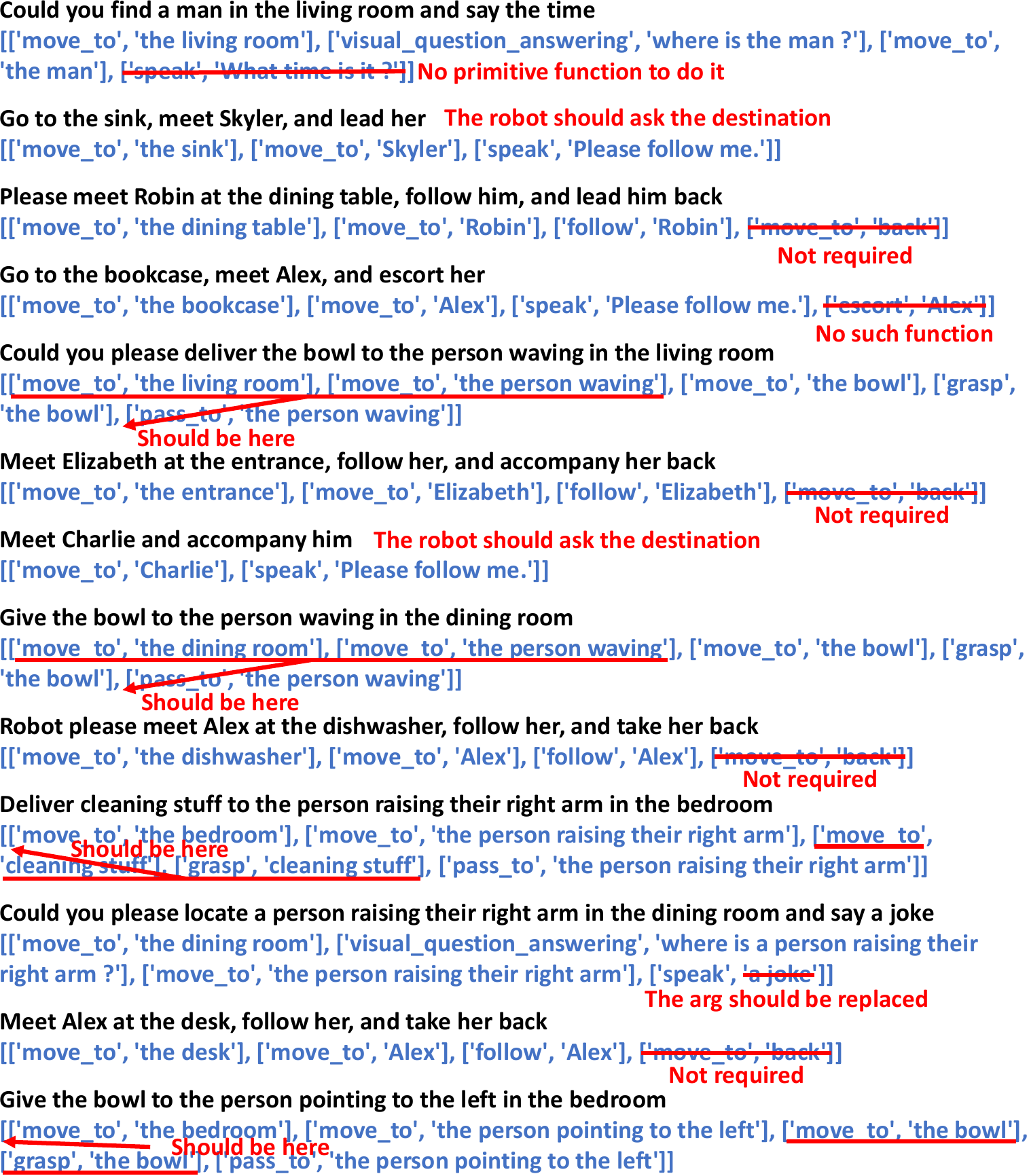}
  \caption{Commands generated by RoboCup@Home Command Generator that incorrectly generate primitive function arguments or order. The black text shows the generated commands, the blue text shows the arguments and order of the generated primitive functions, and the red text shows the parts to be modified and their contents.}
  \label{figure:gpsr-eval}
\end{figure}
\begin{figure}[tb]
  \centering
  \includegraphics[width=\linewidth]{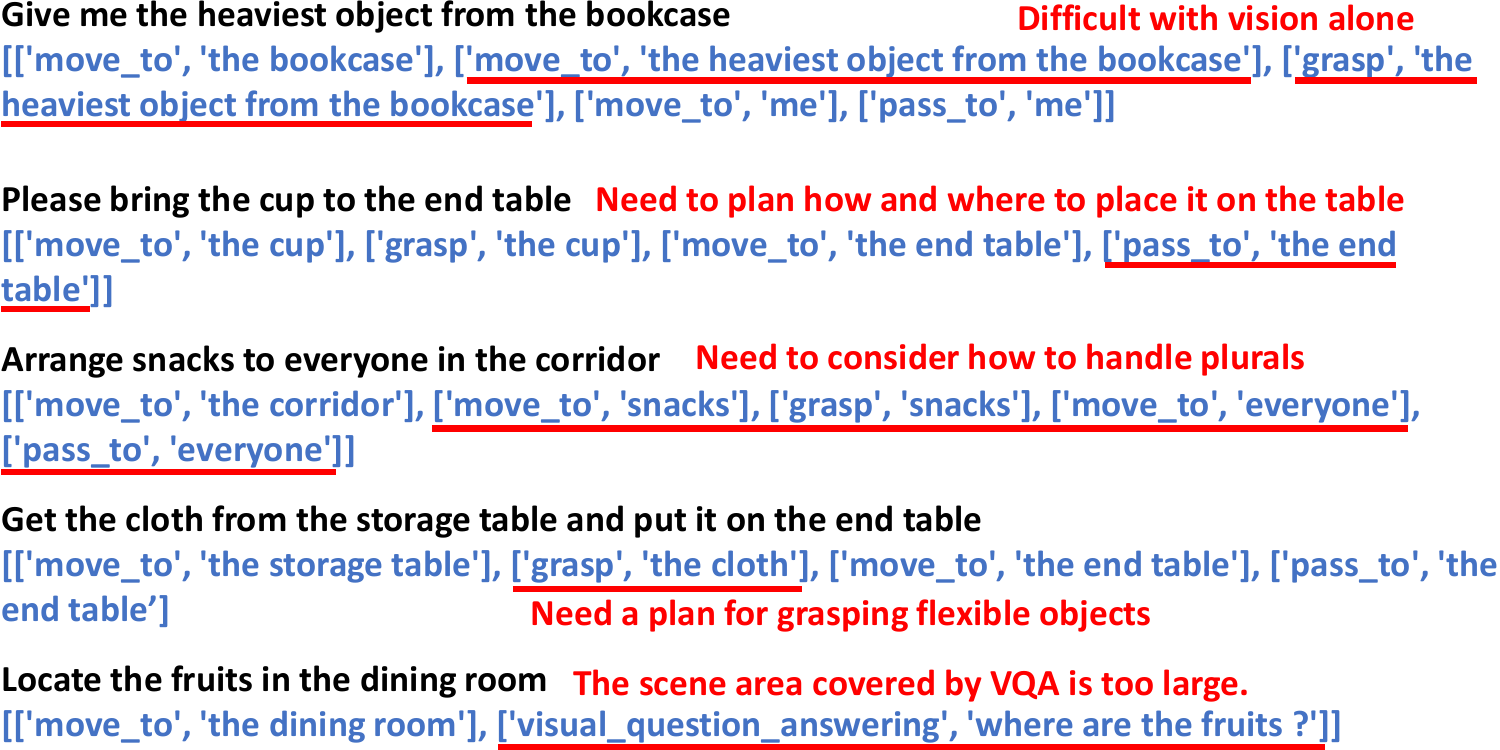}
  \caption{Examples of commands generated by RoboCup@Home Command Generator that correctly generate primitive function arguments and order but are considered challenging to execute with the current primitive function action alone. The black text shows the generated commands, the blue text shows the arguments and order of the generated primitive functions, and the red text shows the parts to be improved and their contents. For example, heavy objects cannot be seen from the image alone but must be held in the hand. When placing grasped items, planning where and how to place them is necessary. The system is difficult to use in the case of multiple objects or all the target persons. VQA for a large area such as a dining room is difficult with only one image.}
  \label{figure:gpsr-eval-2}
\end{figure}
\begin{figure*}[tb]
  \centering
  \includegraphics[width=\linewidth]{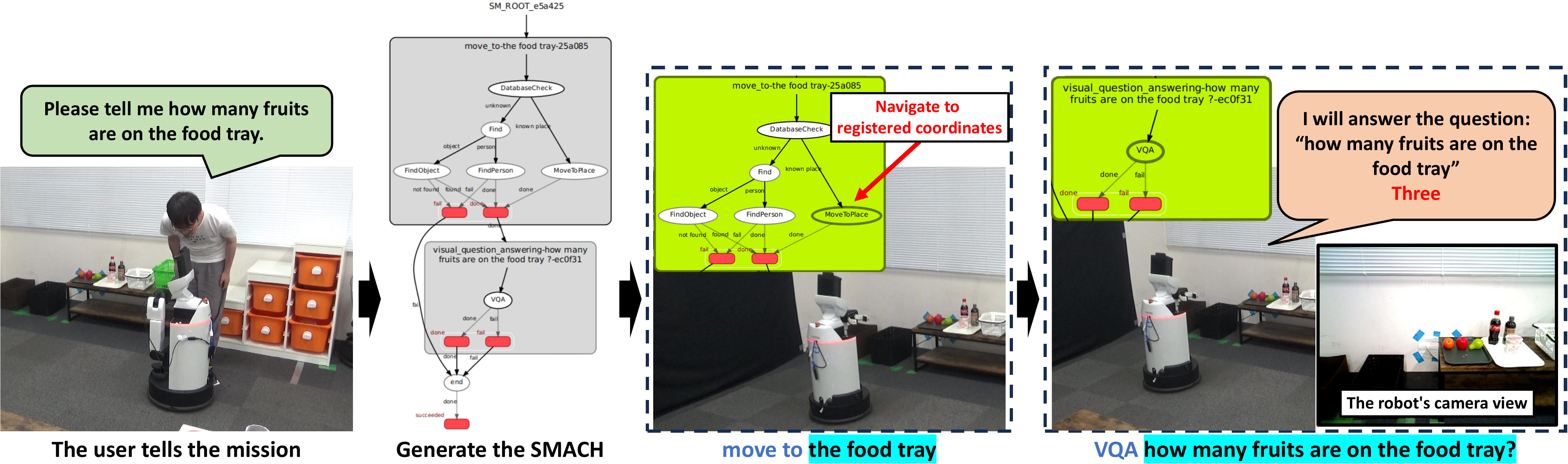}
  \caption{The robot executes the person's command ``Please tell me how many fruits are on the food tray.''. The name of the primitive function being executed is blue-colored, and its arguments are marked in light blue. The robot generated a SMACH, \textcolor{blue}{move\_to the food tray}, \textcolor{blue}{visual\_question\_answering how many fruits are on the food tray}. Then, the food tray is the registered navigation point, so the robot navigates to it. After arrival, the robot entered ``how many fruits are on the food tray'' into the VQA model and obtained the output ``Three''. The robot spoke the result of it and completed the task.}
  \label{figure:exp-vqa}
\end{figure*}
\begin{figure*}[tb]
  \centering
  \includegraphics[width=\linewidth]{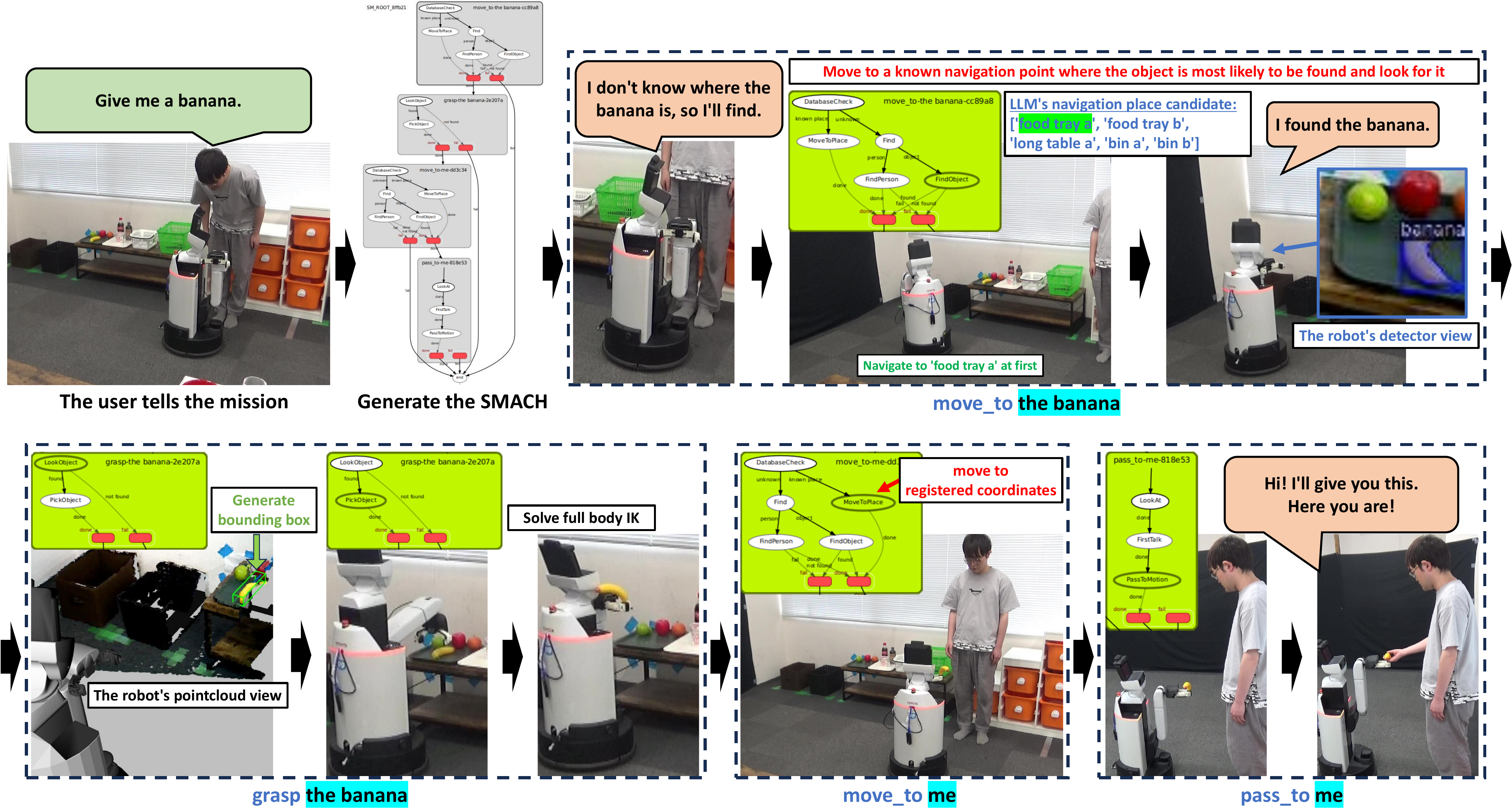}
  \caption{The robot executes the person's command ``Give me a banana''. The name of the primitive function being executed is blue-colored, and its arguments are marked in light blue. The robot generated a SMACH, \textcolor{blue}{move\_to the banana}, \textcolor{blue}{grasp the banana}, \textcolor{blue}{move\_to me}, \textcolor{blue}{pass\_to me}. Then, the banana is not the registered navigation point or a person's name. The robot inferred likely locations by the LLM and visited ``food tray A'', which was registered and most likely to exist. The robot found the ``banana'' from the results of the object detection model, so the \textcolor{blue}{move\_to the banana} was successful. Next, the robot executed \textcolor{blue}{grasp the banana}, which outputs a bounding box covering the banana from the object detection model results and the point cloud, and solves the whole body IK for grasping it. Then the robot executed \textcolor{blue}{move\_to me}. Then the robot executed \textcolor{blue}{pass\_to me} and completed the task.}
  \label{figure:exp-grasp}
\end{figure*}
\begin{figure}[tb]
  \centering
  \includegraphics[width=\columnwidth]{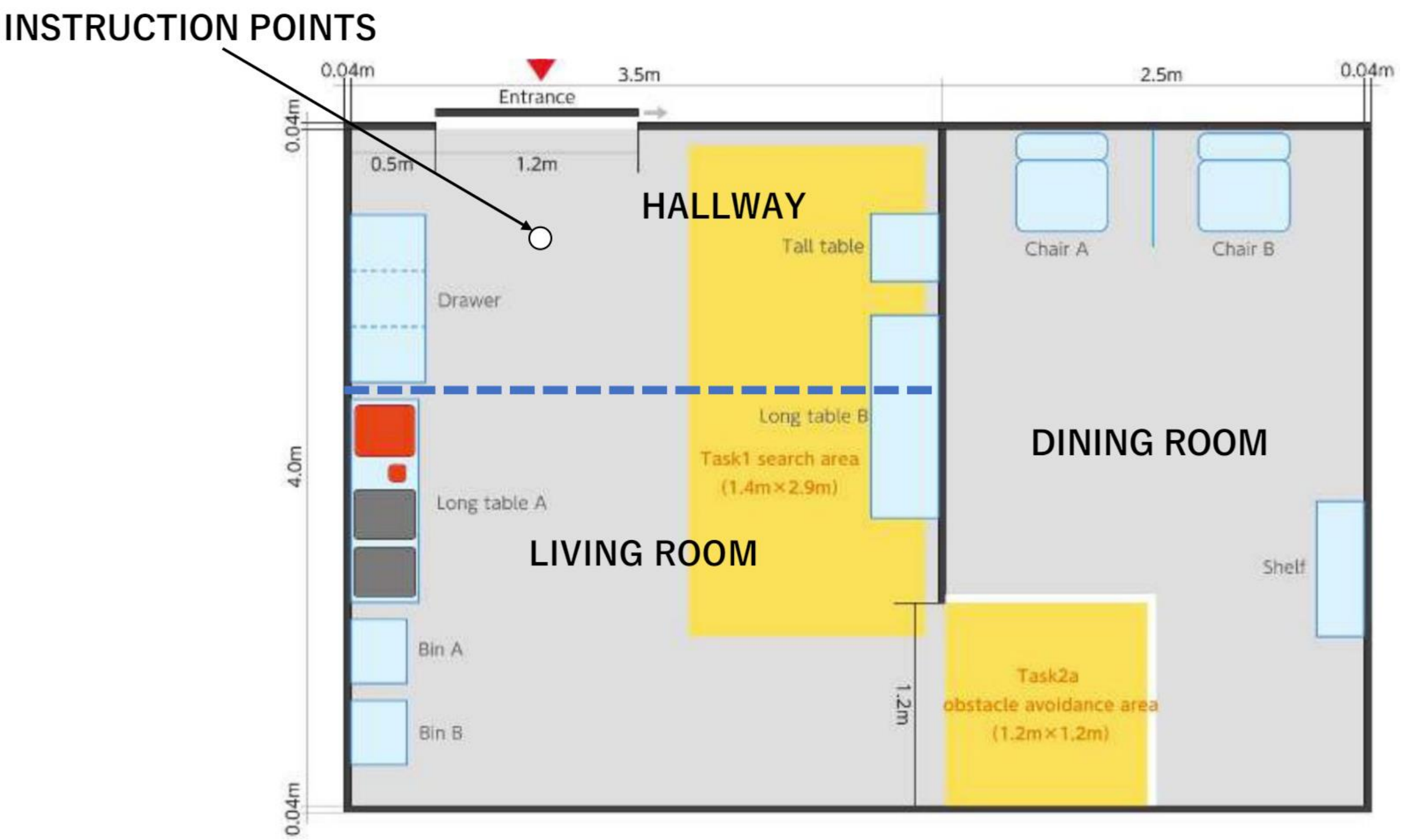}
  \caption{Preannounced map that will be used for the competition \cite{gpsrofficial}. The competition field will be set up based on this map, and each team's robot will prepare a map of its field before the competition.}
  \label{figure:gpsr-map}
\end{figure}
\begin{figure*}[tb]
  \centering
  \includegraphics[width=\linewidth]{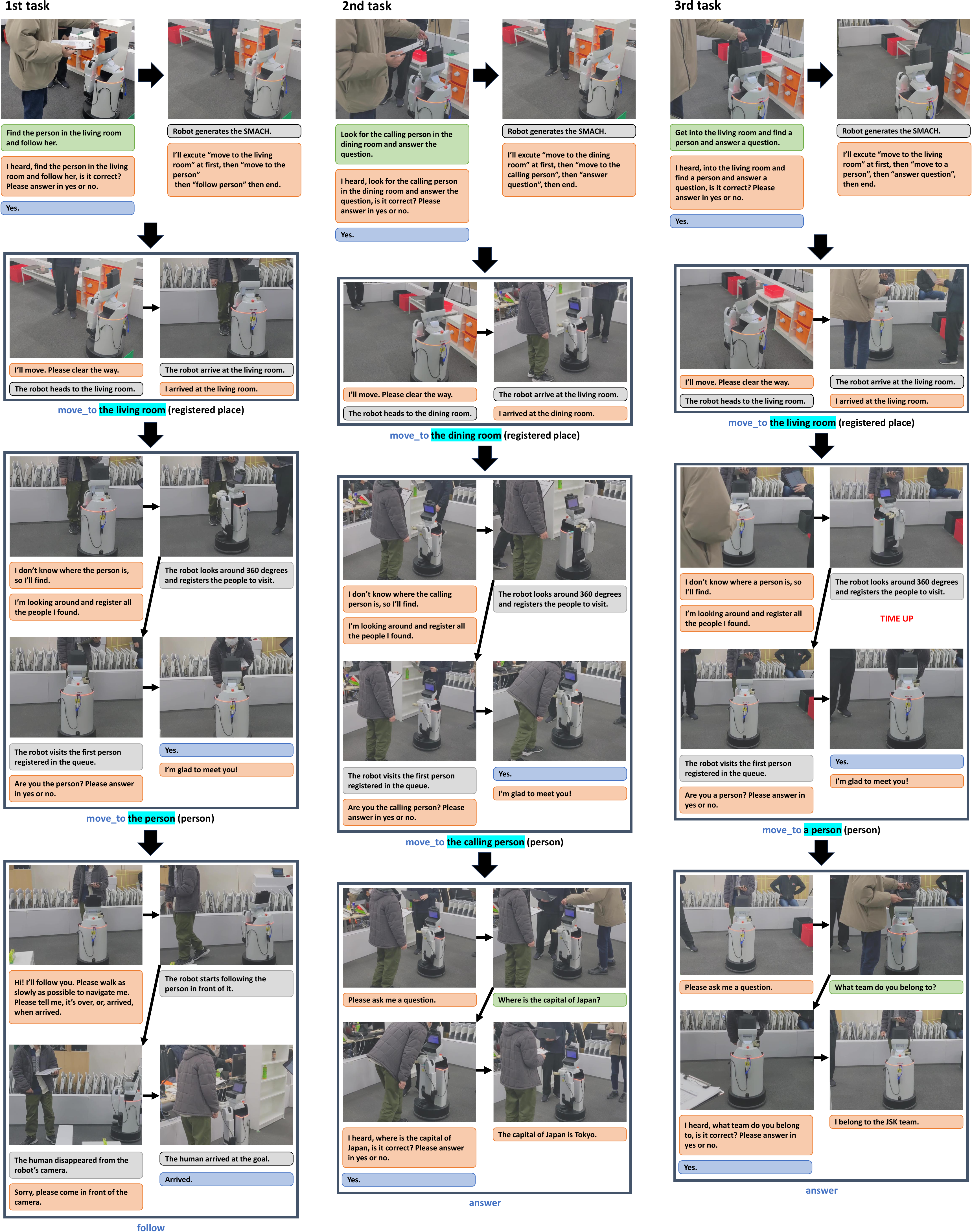}
  \caption{GPSR competition in RoboCup@home 2022 Japan Open. After each task, the robot returns to the Instruction Point to hear the command from the operator. TTS speech is shown in green boxes, robot speech in orange boxes, human speech in blue boxes, and the robot's action in gray boxes. The name of the primitive function being executed is blue-colored, and its arguments are marked in light blue.}
  \label{figure:gpsr-competition}
\end{figure*}

\subsection{Evaluation of Methods to Generate the SMACH by LLM}
\label{expeval}
We evaluated the correct response rate when the command of the RoboCup@Home Command Generator was input into the system. First, we output 100 GPSR commands in sequence from the RoboCup@Home Command Generator and compiled them into primitive functions using the Sec. \ref{gensmach} method. Of these, (1) 87 tasks could correctly output the arguments and order of primitive functions, and (2) 59 of the tasks in (1) might be correctly executed with the current implementation of primitive functions. About (1), we show what commands were wrong in \figref{gpsr-eval}.

We show examples in \figref{gpsr-eval-2} of those that do not satisfy (2). The heaviest object cannot be seen from the image alone. When placing grasping objects in the \textcolor{blue}{pass\_to} function, it is necessary to plan where and how to place them. If the function's arguments are multiple objects or all the target persons, etc., it is difficult for this system to execute the task. VQA for a large area such as a dining room is difficult with only one image.

\subsection{GPSR system experiments with perception task}
We created an environment in our lab for the \figref{gpsr-map} that will be used on the convention day. We named the two trays on Long Table A "Food Tray A" and "Food Tray B" to indicate that food is placed on them and registered their navigation locations to the robot.

\subsubsection{Task Including VQA}
We conducted an experiment in which a human command, ``Please tell me how many fruits are on the food tray.'', let the robot to execute the task of counting and teaching the number of fruits on the food tray. We show the experiment in the \figref{exp-vqa}.

The person told the robot ``Please tell me how many fruits are on the food tray''. Then the robot generated a SMACH based on the output of the LLM, \textcolor{blue}{move\_to the food tray} and \textcolor{blue}{vqa how many fruits are on the food tray?}. Based on the generated SMACH, the robot started to move. First ``food tray'' was the registered navigation point, so the robot navigated to it. Then the robot input the current image and ``how many fruits are on the food tray'' to the VQA model, and it output ``three'', so the robot said ``I will answer the question: how many fruits are on the food tray. Three''. Three fruits were placed on the food tray, and the robot could execute the task correctly.

\subsubsection{Task Including Grasp}
We conducted an experiment in which a human command, ``Give me a banana'', let the robot to execute the task of taking the banana and passing it to the person. We show the experiment in the \figref{exp-grasp}.

The person told the robot, ``Give me a banana''. Then the robot generated a SMACH based on the output of the LLM, \textcolor{blue}{move\_to the banana}, \textcolor{blue}{grasp the banana}, \textcolor{blue}{move\_to me}, \textcolor{blue}{pass\_to me}. Based on the generated SMACH, the robot started to move. First ``the banana'' is not the person's name and registered navigation point, so the robot inferred where it was likely to be on the LLM. The LLM inferred that the banana would exist at ``food tray A'', ``food tray B'', ``long table A'', ``bin A'', and ``bin B'' in that order. The robot first navigated to ``food tray A''. Since the robot detected the banana by the object detection model, it exited the \textcolor{blue}{move\_to the banana} and started \textcolor{blue}{grasp the banana}. The robot output a bounding box covering a banana based on the inference results of the object detection model and the point cloud sensor data and solved the full body IK and grasped it. Then, the robot started \textcolor{blue}{move\_to me}. Since ``me'' was a registered navigation point, the robot navigated to it and executed \textcolor{blue}{pass\_to}, passing the banana to the person. The robot could execute the task correctly.

\subsection{GPSR in robocup@home 2022 Japan Open}
We ran the proposed system in the GPSR competition of the robotcup@home 2022 Japan Open held at The University of Tokyo, Japan, on March 7, 2023.

About the conditions of the competition, the questions the robot would be asked, and the map on which the robot would operate were announced before the competition.

We were told that the following questions are used.
\begin{quote}
  How many days is a week? \\
  What is your name? \\
  Where is the capital of Japan? \\
  Who is the Prime Minister of Japan? \\
  What team do you belong to? \\
  Where will the next Summer Olympics be held? \\
  How many teams are participating in the Domestic Standard Platform League? \\
  What is the highest mountain in the world? \\
  Who wrote Hamlet? \\
  How many hours a day? \\
\end{quote}

We input the knowledge of these questions into the LLM, especially the topical or environment-dependent questions, as shown in Sec. \ref{answer}.

We were told that the navigation point in \figref{gpsr-map} is used. Before the competition begins, teams can make maps in the set-up environment.

Commands from the operator were told to the robot by a TTS (Text to Speech) synthesized voice prepared by the convention.

We show the competition's picture, dialogue, and situation in the \figref{gpsr-competition}.

In the 1st task, the operator (TTS) said, ``Find the person in the living room and follow her''. The robot said it would execute \textcolor{blue}{move\_to the living room}, \textcolor{blue}{move\_to the person}, \textcolor{blue}{follow person} in order. Next, the robot navigated to the living room since it was a registered navigation point. Next, the robot does not know where ``the person'' is, and the LLM inferred ``the person'' as a person, so it started to find the person. The robot rotated 360 degrees on the spot, registering the people to visit by the human skeleton estimation model and then visiting the first person. The robot asked him ``Are you the person? Please answer in yes or no.'', then he answered ``Yes'', so the robot finished \textcolor{blue}{move\_to the person} function and started \textcolor{blue}{follow} function. He started to navigate to the Dining Room. While following, the robot encouraged him to come in front of the camera by saying ``Sorry, please come in front of the camera'' when he went off camera. When he reached his destination, he said ``Arrived,'' so the robot finished the \textcolor{blue}{follow} and returned to the Instruction Point.

In the 2nd task, the operator (TTS) said ``Look for the calling person in the dining room and answer the question''. The robot said it would execute \textcolor{blue}{move\_to the dining room}, \textcolor{blue}{move\_to the calling person}, \textcolor{blue}{answer question} in order. Next, the robot navigated to the Dining Room sice it was a registered navigation point. Next the robot does not know where ``the calling person'' is, and the LLM inferred ``the calling person'' as a person, so it started to find the person to visit like the 1st task. Next, the robot started to execute the function \textcolor{blue}{answer}, and let him to say a question. The TTS said ``Where is the capital of Japan'' and the robot repeated it and asked whether it listened is correct or not. He said ``Yes'' so the robot input the question to the LLM and said its output ``The capital of Japan is Tokyo''. Then the robot finished \textcolor{blue}{answer} function and retured to the Instruction Point.

In the 3rd task, the operator (TTS) said, ``Get into the living room and find a person and answer a question''. The robot said it would execute \textcolor{blue}{move\_to the living room}, \textcolor{blue}{move\_to a person}, \textcolor{blue}{answer question} in order. Like the 1st task and 2nd task, first, the robot navigated to the Living Room and visited the person to visit. Next, the robot started \textcolor{blue}{answer} and asked ``What team do you belong to?'', then said the LLM's output, ``I belong to the JSK team''. The robot finished the \textcolor{blue}{answer} function and returned to the Instruction Point. In the 3rd task, 10 minutes had passed from the start of the competition, but the task was performed until the end of the robot action with the competition's consideration.

\tabref{gpsr-result} shows the competition score. Using the proposed system, our team won first place.

\begin{table}[h]
  \caption{The score result in GPSR}
  \label{table:gpsr-result}
  \begin{center}
    \begin{tabular}{|c|c|c|}
      \hline
      Rank & Team name & Score \\
      \hline
      1 & \bf{Team JSK (ours)} & \bf{130} \\
      2 & TRAIL & 41.25 \\
      3 & Hibikino-Musashi@Home & 25 \\
      4 & eR@sers & 12.5 \\
      4 & OIT-RITS & 12.5 \\
      4 & \begin{CJK}{UTF8}{ipxm}あばうたぁ～ず\end{CJK} & 12.5 \\
      7 & SOBITS & 0 \\
      \hline
    \end{tabular}
  \end{center}
\end{table}

\section{Discussion}
\label{discussion}

\subsection{Primitive Function Generation}
For a system that generates a sequence of primitive functions from natural language, Sec. \ref{expeval} shows that a high percentage of correct primitive function sequences are generated. About the incorrect outputs, we can consider adding new examples to the LLM input. If there is missing information in the commands, it is difficult for the system to compensate for it on its own. Since the robot first confirms a sequence of primitive functions to be executed by the person, the person can correct or make the command more concrete.

\subsection{\textcolor{blue}{visual\_question\_answering}}
The VQA in this study lets a robot navigate and face a predetermined direction, then captures images of the robot's camera. However, in a real daily life environment, it is difficult for the robot to answer a question like ``How many specific foods are in the kitchen'' or ``How many specific consumables'' with a single image. The robot needs to search around the environment to answer these questions. As a solution to it, there is a method called ScanQA \cite{azuma2022scanqa}, which can perform Question Answers based on the 3DScan information of the environment. When we want to use it in a real daily life environment, the robot has to plan the range to be scanned, its movement path, camera trajectory, and the manipulation of doors, drawers, cupboards, etc., using manipulation.

\subsection{\textcolor{blue}{pass\_to}}
In this research, \textcolor{blue}{pass\_to} is a primitive function that releases the object in the robot's hand to pass/place it to/at the argument. To place an object in a narrow space, the robot has to know the pose of the grasping object, reposition it, and re-grasp it if needed, then place \cite{wada2022reorientbot}. If there is no temporary space to place, there is an option to use a dual-armed robot and reposition the object by itself.

\subsection{Continuing the Task even if the Function Fails}
For example, in our implementation, if \textcolor{blue}{move\_to(ARG)}'s argument object is not found, \textcolor{blue}{move\_to} fails, and the robot returns to the Instruction Point. However, we want the robot to search for objects in a real daily life environment persistently. We can create an action such as \textcolor{blue}{ask\_to\_person}, ask the person what to do next when \textcolor{blue}{move\_to} fails. To achieve this in a real daily life environment, we think the implementation that generates additional SMACHs from the information heard from the person in \textcolor{blue}{ask\_to\_person}.

\subsection{Local Knowledge}
The output of the LLM is based on the general knowledge of the world as described in Sec. \ref{introduction}. However, in a real daily life environment, there is much knowledge specific to the environment, such as the meaning of places on a map, the arrangement of furniture and objects, and to whom objects belong. In this study, we input prior information into the LLM in language, as shown in Sec.\ref{moveto} and Sec.\ref{answer}. There is research to collect object data of the environment on a daily basis \cite{furuta2018everyday}. If we can assign multiple short word tags to these data, the LLM can process local knowledge like Sec.\ref{moveto} and Sec.\ref{answer}.

\subsection{Detection and Improvement of Primitive Functions Failures}
Each primitive function has two types of failures: programmatic failures and failures where the program terminates successfully, but the result is different. In addition, for example, \textcolor{blue}{grasp} has a failure where the robot falls off the hand while executing another function, even though \textcolor{blue}{grasp} succeeded at that time. It is difficult for a robot to detect these failures by itself, so we can think of developing a feedback system. We can consider the user interface in which the robot stores data during the execution of each function and receives feedback from a person after completing a task based on the collected data.

\section{Conclusions}
\label{conclusion}
In this study, we proposed a system to realize a general purpose service robot using the foundation model. This system prepares seven primitive functions of the robot in advance. The system compiles the commands in spoken language into the primitive function sequence by the LLM and generates the state machine. Next, for each primitive function, the vocabulary of the VLM is automatically changed. The LLM further instantiates the action if the function's argument is ambiguous. The state machine task executable manages all actions, and the following actions upon success or failure of each action are all deterministic. We found this method helpful through experiments with actual robots and competition evaluations. At the same time, we found imperfections in each primitive function, problems with local knowledge, and challenges in detecting and learning from failures. We believe that the imperfections of each primitive function will be improved by the development of computer vision and planning research, the verbalization of collected data will improve the problem of local knowledge, and failure detection and learning will be improved by the development of user interface and robot behavior learning.

Finally, research on daily life support robots using the foundation model will spread widely and rapidly. We hope this paper will help configure such systems and discover issues to be addressed.

\addtolength{\textheight}{-12cm}   

\bibliographystyle{junsrt}
\bibliography{main}

\end{document}